\documentclass[numsec,webpdf,modern,medium,namedate,arxiv]{oup-authoring-template}

\onecolumn

\graphicspath{{figures/}}

\theoremstyle{thmstyleone}%
\theoremstyle{thmstyletwo}%
\theoremstyle{thmstylethree}%

\usepackage[fontsize=12pt]{fontsize}

\newcommand{\st}{\operatorname{s.t.}}
\newcommand{\argmin}{\operatornamewithlimits{argmin}}
\newcommand{\M}{\mathcal{M}}
\newcommand{\y}{\mathbf{y}}
\newcommand{\wy}{\widehat{\y}}
\newcommand{\Xm}{\mathbf{X}}
\newcommand{\Sm}{\mathbf{S}}
\newcommand{\err}{\boldsymbol{\varepsilon}}
\newcommand{\Ell}{\mathcal{L}}
\newcommand{\Prob}{\operatorname{P}}
\newcommand{\E}{\operatorname{E}}

\newcommand{\U}{\operatorname{\delta}}
\newcommand{\Um}{\hat{\mathbf{U}}}
\newcommand{\given}{\mid}
\newcommand{\am}{\boldsymbol{\alpha}}
\newcommand{\bm}{\boldsymbol{\beta}}

\begin{document}

\journaltitle{Journals of the Royal Statistical Society}
\DOI{DOI HERE}
\copyrightyear{2025}
\pubyear{XXXX}
\access{Advance Access Publication Date: Day Month Year}
\appnotes{Original article}

\firstpage{1}

\title[Uncertainty in Fair Machine Learning]
  {The Role of Uncertainty in Assessing the Fairness of Machine Learning Models}

\author[1,2]{Francesca Panero}
\author[3,$\ast$]{Ernst C. Wit}
\author[4]{Marco Scutari}

\authormark{Panero et al.}

\address[1]{
  \orgdiv{Department of Methods and Models for Economics, Territory and Finance},
  \orgname{Sapienza Universit\`a di Roma},
  \orgaddress{\street{Rome}, \postcode{00161}, \country{Italy}}
}
\address[2]{
  \orgdiv{Department of Statistics},
  \orgname{London School of Economics and Political Science},
  \orgaddress{\street{London}, \postcode{WC2B 4RR}, \country{UK}}
}
\address[3]{
  \orgdiv{Faculty of Informatics},
  \orgname{Universit\`a della Svizzera Italiana (USI)},
  \orgaddress{\street{Lugano}, \postcode{6962}, \country{Switzerland}}
}
\address[3]{
  \orgname{Istituto Dalle Molle di Studi sull'Intelligenza Artificiale (IDSIA)},
  \orgaddress{\street{Lugano}, \postcode{6962}, \country{Switzerland}}
}

\corresp[$\ast$]{
  Address for correspondence. Ernst Wit, Faculty of Informatics, Universit\`a della Svizzera Italiana (USI), Lugano, 6962, Switzerland. \href{email:ernst.jan.camiel.wit@usi.ch}{ernst.jan.camiel.wit@usi.ch}
}

\abstract{
Machine learning models are widely used in clinical applications, social media, law enforcement and critical infrastructure. Verifying whether their outputs are biased against disadvantaged groups or individuals is crucial to ensuring they are fair and allowing their use in such settings. A rigorous risk assessment of possible fairness violations requires quantifying the uncertainty associated with selecting and estimating such models. Yet, this is rarely done in the literature, which focuses on identifying a single model with a suitable trade-off between predictive accuracy and fairness. In this paper, we move beyond point estimation and discuss frequentist and Bayesian approaches to uncertainty quantification for fair machine learning, with practical examples and implications for simulated and real data.
}
\keywords{fairness, machine learning, uncertainty quantification, bootstrap, Bayesian inference, risk assessment}

\maketitle

\section{Introduction}
\label{sec:intro}

Assessing the \emph{fairness} of machine learning models is the process of identifying and eliminating algorithmic bias against individuals belonging to disadvantaged groups. Tools and commercial products based on such models, often called ``AI systems,''  are required to be ``developed and used in a way that includes diverse actors and promotes equal access, gender equality and cultural diversity, while avoiding discriminatory impacts and unfair biases that are prohibited by Union or national law'' by the EU AI Act \citep{eu-ai}. The US \citep{ai-act} and Japan's AI Act \citep{jp-ai} contain similar provisions. However, these legal frameworks and the associated guidelines also recognise that the possibility of unfair model outputs cannot be excluded entirely without impacting the models' suitability for practical use. For instance, the NIST AI Risk Management Framework \citep{nist-ai} explicitly states we should aim to ``minimise anticipated negative impacts of AI systems and identify opportunities to maximise positive impacts,'' discussing how ``trade-offs may emerge between optimising for interpretability and achieving privacy'' and ``privacy-enhancing techniques can result in a loss in accuracy, affecting decisions about fairness'' in different domains and scenarios. The trade-off between fairness and predictive accuracy is also widely documented in the literature \citep{menon,zhao}, and overcoming it remains an active research topic \citep[e.g.,][]{wick}.

To fulfil these legal obligations, we must assess the risks associated with machine learning models by weighing the impact of various types of algorithmic bias against the probability of observing them. In turn, this requires us to \emph{quantify the uncertainty} in the model outputs, including that arising from model selection, parameter estimation and the inherent stochastic noise in both training and target data. How to do that is not well explored in the literature, which mainly frames fairness as a constrained optimisation problem that produces only point parameter estimates \citep{pessach}. In particular, the trade-off between accuracy and fairness is almost exclusively formulated as optimising for goodness of fit while constraining some mathematical definition of fairness using for example techniques from convex \citep{zafar}, non-convex \citep{komiyama}, graph \citep{agarwal} or functional optimisation \citep{pontil}. Recently, attempts to introduce uncertainty in fairness have looked at how to reduce the impact of lack of information in data, especially regarding disadvantaged categories \citep{lee2026fairness, singh2021fairness, zhang2023individual}, and how to embed uncertainty in the fairness definition \citep{tahir2023fairness, romano2020malice}. The landscape, though, is still dominated by point predictions, largely because the machine learning and computer science culture of algorithmic fairness prioritises prediction accuracy over accurate modelling \citep{breiman2001statistical, shmueli2010explain}.

This paper highlights the implications of uncertainty quantification in fair machine learning and offers suggestions for implementing it using frequentist and Bayesian approaches. Unlike to previous work in fairness uncertainty, we provide a characterisation of uncertainty leveraging a more classical statistical toolbox, which allows us to have exact rigorous results with well-understood properties, look at the uncertainty quantification problem from multiple points of view, and provide different solutions that can be useful for stakeholders in fairness. 
The approaches we describe, bootstrap resampling and hierarchical modelling, are representative examples of what frequentist and Bayesian statistics make possible in this setting; they are not meant as normative or universal recommendations. How we ensure fairness depends on which mathematical definition of fairness is most appropriate in a specific application; many are reviewed in \citet{mehrabi} and \citet{pessach}. It also depends on whether we are enforcing fairness by transforming the data before we analyse them (``pre-processing''), by appropriately selecting and estimating the machine learning model (``in-processing''), or by adjusting model predictions afterwards (``post-processing''). The literature presents several examples of each approach \citep[][among others]{jieng,chzhen,frappe}, but again compares them mainly in terms of accuracy. How much fairness we choose to enforce, different fairness definitions and pre-, in- and post-processing approaches will produce conclusions with different amounts of uncertainty, which should be taken into account when assessing model risk, as we argue in the following.

The remainder of the paper is structured as follows. We will introduce some notation and review the literature in Section~\ref{sec:notation}. In Section~\ref{sec:uncertainty}, we formulate uncertainty quantification in the context of fair machine learning, noting the limitations of pre- and post-processing approaches. We then describe frequentist and Bayesian approaches to uncertainty quantification for in-processing approaches in Section~\ref{sec:uq}. In Section~\ref{sec:results}, we first illustrate our findings on simulated data, investigating how correlation between sensitive and non-sensitive predictors and different fairness values affect uncertainty. Secondly, we apply the methodologies to two benchmark real data sets and discuss the implications of our findings.

\subsection{Background and notation}
\label{sec:notation}

Let $\Xm$ be a set of predictors, $\Sm$ a set of sensitive attributes (disjoint from $\Xm$), and $\y$ a response variable. We define a machine learning model $\M$ as
\begin{equation*}
  \y = f(\Xm, \Sm; \theta) + \err,
\end{equation*}
where $f(\cdot)$ is a generic function with parameters $\theta$ mapping $\Xm$ and $\Sm$ to $\y$, and $\err$ is an error term
with mean zero. The sensitive attributes represent variables, such as gender and race, that may have a statistical association with $\y$ but should not be used, or should be limited in their predictive power, to estimate $\y$ via $\wy = f(\Xm, \Sm; \widehat{\theta})$  to avoid disparate (unfair) treatment of some of the groups or individuals they identify. 

Given data $\{\y,\Xm,\Sm\}$, learning $\M$ involves selecting the optimal $f(\cdot)$ by minimising a loss function $\Ell(\y, f(\Xm, \Sm; \theta))$ with respect to $\theta$. Unless we restrict learning in some way, the estimated values $\wy$ may violate some form of \emph{fairness}, because they are a function of the sensitive attributes $\Sm$.

The literature has proposed several approaches to mitigate this problem. \emph{In-processing approaches} to fairness, such as \citet{zafar}, \citet{chzhen} and \citet{stco21}, formulate learning as
the constrained optimisation problem
\begin{equation*}
  \hat{\theta}_{\mathit{fair}}(r) = \argmin_{\theta\in\Theta} \Ell(\y,f(\Xm, \Sm; \theta)) \quad \st \quad \Ell_\mathit{fair}(\y, f(\Xm, \Sm; \theta), \phi) \leqslant r,
\end{equation*}
where $r\ge 0$ is a user-defined unfairness level and $\phi$ are optional parameters of some \emph{fairness loss} $\Ell_\mathit{fair}$, which quantifies the amount of unfairness arising from using $\Sm$ in $\wy$, attributed to $\M$ by extension.
Equivalently, the problem can be stated as the penalised loss minimisation
\begin{equation*}
  \hat{\theta}_{\mathit{fair}}(r) = \argmin_{\theta\in\Theta} \Ell(\y,f(\Xm, \Sm; \theta)) + \lambda(r) \Ell_\mathit{fair}(\y, f(\Xm, \Sm; \theta), \phi),
\end{equation*}
commonly studied for penalised regression models \citep{elemstatlearn}. For brevity, in the rest of the paper we will use $\lambda$ in lieu of $\lambda(r)$.

Several proposals for the fairness loss can be found in the literature. If fairness is defined as statistical parity (also known as demographic parity) in a classification problem with one binary sensitive variable, the constraining function $\Ell_\mathit{fair}(\y, f(\Xm, \Sm; \theta), \phi)$ can be $|\mathbb{P}(\mathbf{Y}=\y|S=1)-\mathbb{P}(\mathbf{Y}=\y|S=0)|\le r$ for all $\y$. In this case, $\Ell_\mathit{fair}(\y, f(\Xm, \Sm; \theta), \phi)$ is constant in $\y$ and does not depend on other parameters $\phi$. Other fairness definitions condition instead on the true value of $\y$. For example, equalised odds require that protected and unprotected groups are offered similar rates of classification errors, as measured by false-positive and false-negative rates.

\emph{Post-processing approaches} to fairness, such as that from \citet{frappe}, assume that the model $\M$ is given and evaluate $\Ell_\mathit{fair}(\cdot)$ on the predicted values $\wy$ produced by $\M$ for a validation set. Then one selects the transformation $T_\lambda^*(\cdot)$ of the data that balances the predictive accuracy of $T_\lambda(\wy)$ with its fairness as defined by $\Ell_\mathit{fair}(\cdot)$, i.e.,
\begin{equation*}
  T^*_\lambda = \argmin_{T \in \mathcal{T}} \Ell(T(\wy) - \y)) - \lambda \Ell_\mathit{fair}(T(\wy)).
\end{equation*}

\emph{Pre-processing approaches} to fairness do not have access to $\M$, nor to its estimation process. Instead, they typically replace it with an elementary statistical model, such as a simple linear regression or a shallow classification tree, and transform the original $\y$ and $\Xm$ so that $\Ell_\mathit{fair}(\wy, \Sm; \theta) \leqslant r$. 

These approaches share a tuning parameter $\lambda$ or $r$ that controls fairness, trading off accuracy in the process. Hence, model selection must choose this and other tuning parameters to achieve a suitable balance between these two properties.

\subsection{Implications of pre-, in- and post-processing  approaches in uncertainty quantification}
\label{sec:uncertainty}

The predicted values $\wy$ carry the uncertainty arising from multiple sources for all three classes of fair machine learning approaches. 
Without loss of generality, we follow the notation in \citet{berger} and write the predictive densities of $\wy$ by marginalising over parameters, models and data in turn:
\begin{align}\label{eq:densities}
  p(\wy \given \M, D) &= \int p(\wy \given \M, \theta, D)\,
     \Prob(d\theta \given \M, D), \notag\\
  p(\wy \given D) &= \int p(\wy \given \M, D)\, \Prob(d\M \given D),\notag\\
  p(\wy) &= \E_D\big[ p(\wy \given D) \big],
\end{align}
where $p(\wy \given \M, \theta, D)$ is the predictive distribution of a new data point given the model, the parameters and the data; $\Prob(\theta \given \M, D)$ is the posterior density of the model's parameters; $\Prob(\M \given D)$ is the posterior probability of the model; and the last expectation is taken with respect to the unknown data distribution.
All densities above are conditional on the fairness constraint $\lambda$, which we assume to be fixed by the analyst, and we suppress it from the notation for readability. 

In a machine learning context, we are mainly interested in predictive uncertainty \citep[see][for a review]{tyralis}. Let $\U(\cdot)$ be any measure of the uncertainty in a predictive distribution that is concave, in the sense that a mixture of distributions is at least as uncertain as the average of its components. The variance and the entropy both satisfy this condition, as does the Bayes risk of any loss function \citep{degroot}. If we consider the three levels of conditioning in turn, we can decompose the uncertainty:
\begin{align}
  \U(\wy) ={}
  & \underbrace{\E_{D,\M,\theta}\big[\U(\wy \given \M,\theta,D)\big]}_{\text{aleatoric}} \notag\\
  &+ \underbrace{\E_{D,\M}\big[\U(\wy \given \M,D)
       - \E_{\theta \given \M,D} \U(\wy \given \M,\theta,D)\big]}_{\text{model estimation}} \notag\\
  &+ \underbrace{\E_{D}\big[\U(\wy \given D)
       - \E_{\M \given D} \U(\wy \given \M,D)\big]}_{\text{model selection}} \notag\\
  &+ \underbrace{\U(\wy) - \E_{D}\U(\wy \given D)}_{\text{data}}.
\label{eq:uncertainty}
\end{align}
where the expectations are taken with respect to the conditional densities introduced in Eq.~\ref{eq:densities} and the corresponding unconditional ones.
The aleatoric term represents the uncertainty that remains no matter how much we learn about the model and its parameters. The remaining terms make up the epistemic uncertainty: the lack of knowledge that can be reduced by improving our estimation, model selection and data quantity. It is composed of the model estimation term, telling us what we would gain by learning $\theta$, having fixed $\M$ and $D$; the model selection term, the information we would gain by learning the model, having observed $D$; and finally the sampling variability term, the knowledge produced by observing the sample itself. Taking $\U(\cdot)$ to be the variance, the expression above reduces to the law of total variance applied at each level.

Ideally, we want to reduce $\U(\wy)$ to the point where we can tell with a set
confidence level that $\M$ is unlikely to produce an unfair $\wy$. 
This is impaired with a post-processing methodology, since it is impossible to accurately estimate $\U(\wy)$ in
those cases: $\M$ is considered a black box, we cannot
quantify the uncertainty arising from either model selection or model estimation, while we can only measure the uncertainty associated with
selecting $T(\cdot)$ and the inherent stochasticity of $\wy$. Pre- and
post-processing approaches also introduce auxiliary functions and models that
require their own tuning, which adds uncertainty to the overall model selection
process. Both perform this tuning independently of the selection and estimation process of $\M$, making the calibration of the overall process difficult.
Therefore, accurate uncertainty quantification, and thus risk assessment, is more accurately done with in-processing.

The decomposition above applies unchanged to the fairness loss $\Ell_\mathit{fair}$, whose distribution is induced by the same hierarchy. 
It is the uncertainty in $\Ell_\mathit{fair}$, rather than that in $\wy$, that measures the risk of deploying an unfair model. In our work, we address how to estimate part of the epistemic uncertainty (model estimation and sampling variability) and the uncertainty of $\Ell_\mathit{fair}$. As we study non-Bayesian approaches, Eq.~\ref{eq:uncertainty} does not strictly apply as written; even so, we refer to it as it broadly reflects typical machine learning model structural hyperparameter tuning (model selection) and learning (model estimation) and can provide a framework to characterise their variability, for instance, as captured in different bootstrap nesting levels in Section~\ref{subsec:freq_uncertainty}.

\section{Methods}
\label{sec:uq}

In the following sections, we introduce some methodologies for evaluating uncertainty in algorithmic fairness. In Section~\ref{subsec:freq_uncertainty}, we use bootstrap approaches to study variability in model estimation and fairness due to the observed data, construct confidence intervals for the model parameters, and examine how unfairness and accuracy vary as the sample varies. Such approaches are general and can be applied to any model $\M$. Using Bayesian tools, in Section~\ref{subsec:bayesian_uncertainty} we focus on measuring uncertainty from the model-estimation term in the decomposition of Eq.~\ref{eq:uncertainty} by providing credible intervals for the parameters and studying the posterior of the fairness measure. We do so using the model proposed by \citet{stco21}, which allows us to work with linear and generalised linear models, several fairness definitions and multiple sensitive predictors. This lets us work with either closed-form posteriors or Markov Chain Monte Carlo estimates, which, unlike the Bayesian neural networks sometimes used in this literature, provide verifiable estimation and uncertainty quantification. The methodological proposals are presented in this Section, and their applications to simulated and real data in Section~\ref{sec:results}.

\subsection{Frequentist uncertainty quantification of fairness}\label{subsec:freq_uncertainty}

The fair model estimation in Section~\ref{sec:notation} is usually formulated as a penalised likelihood regression problem. This suggests that resampling approaches that assess uncertainty in this class of models \citep{hinkley} are appropriate in this setting. We focus on two tasks: constructing confidence intervals for individual parameters $\theta$ and assessing the risk of fairness violations in $\M$ as a whole.

Nonparametric bootstrap can be used to estimate frequentist confidence intervals for the coefficients of regression models. In models with explicit parameters associated with the effect of individual sensitive attributes, their confidence intervals indirectly assess model fairness: the farther an interval lies from zero, the stronger the effect of the associated sensitive attribute. Conversely, we can argue that if all the confidence intervals of the sensitive attributes include zero, the model is fair, in the sense that the effect of sensitive attributes is not statistically significant. Clearly, for large sample sizes, statistical significance should be balanced with practical significance, as even minor effects can become statistically significant.

In non-penalised linear regression models, resampling the empirical residuals $\widehat{\varepsilon}$ is statistically more efficient than resampling the predictors \citep{legendre}. This advantage is gradually lost in penalised models because $\widehat{\varepsilon}$ becomes increasingly non-orthogonal (and thus non-pivotal) to predictors as the penalty increases. Furthermore, the shrinkage imposed by the fairness penalty is likely to result in artificially narrow confidence intervals. 

Double bootstrap \citep{vinod} builds on Efron's debiased percentiles \citep{efron} to approximately recover pivotality. As the name suggests, it employs a nested bootstrap scheme in which the estimated bias between the first-level and the second-level estimates of $\theta$ is used to correct the empirical quantiles that delimit the confidence interval.
We can use bootstrap resampling to assess the risk that our fair model estimation procedure will yield a model that falls below the required level of fairness when predicting new individuals. With nested bootstrap, we can first average over training-validation splits (to account for drift and sample heterogeneity) and then inside training splits (to account for the variability in parameter estimates) to obtain a marginal estimate of the model fairness, in the following way:

\begin{enumerate}
  \item For $b = 1, \ldots, B_1$:
  \begin{enumerate}
    \item Draw a (nonparametric) bootstrap sample $\y^{(b)}$, $\Xm^{(b)}$ and $\Sm^{(b)}$ from $\y$, $\Xm$, $\Sm$.
    \item Split the bootstrap samples into training ($\y^{(b)}_{\mathrm{TR}}$, $\Xm^{(b)}_{\mathrm{TR}}$ and $\Sm^{(b)}_{\mathrm{TR}}$) and validation sets ($\y^{(b)}_{\mathrm{VA}}$, $\Xm^{(b)}_{\mathrm{VA}}$ and $\Sm^{(b)}_{\mathrm{VA}}$).
    \item For $b' = 1, \ldots, B_2$:
    \begin{enumerate}
      \item Draw a (nonparametric) bootstrap sample $\y^{(b')}_{\mathrm{TR}}$, $\Xm^{(b')}_{\mathrm{TR}}$ and $\Sm^{(b')}_{\mathrm{TR}}$ from $\y^{(b)}_{\mathrm{TR}}$, $\Xm^{(b)}_{\mathrm{TR}}$ and $\Sm^{(b)}_{\mathrm{TR}}$.
      \item Estimate a model $\mathcal{M}^{(b')}$ with the required level of fairness from $\y^{(b')}_{\mathrm{TR}}$, $\Xm^{(b')}_{\mathrm{TR}}$ and $\Sm^{(b')}_{\mathrm{TR}}$.
      \item Estimate the value of the fairness loss for model $\mathcal{M}^{(b')}$ on the validation set ($\y^{(b)}_{\mathrm{VA}}$, $\Xm^{(b)}_{\mathrm{VA}}$ and $\Sm^{(b)}_{\mathrm{VA}}$).
      \item Predict $\y^{(b)}_{\mathrm{VA}}$ from $\Xm^{(b)}_{\mathrm{VA}}$ and $\Sm^{(b)}_{\mathrm{VA}}$ using $\mathcal{M}^{(b')}$ to estimate predictive accuracy.
    \end{enumerate}
  \end{enumerate}
  \item Assess the joint distribution of empirical fairness and predictive accuracy to find the best compromise, either by visual inspection or using a loss function.
\end{enumerate}

In practice, we may choose the pinball loss, or a similar loss from the quantile regression literature \citep{pinball}, which is asymmetric in scoring deviations from the reference value (our required level of fairness). Models that are fairer than required are acceptable, even though we must evaluate the loss of predictive accuracy that comes with the extra fairness. Models that are too unfair should be associated with markedly higher values of the loss function since they are not legally viable for practical use. We may also compute the expectation of the losses over the validation sets to obtain a univariate measure of risk, which can then be used to rank fair model selection and estimation approaches, as well as tuning parameter sets.

\subsection{Bayesian uncertainty quantification of fairness}\label{subsec:bayesian_uncertainty}

In this section, we consider a Bayesian implementation of a regression problem, with the aim of quantifying fairness uncertainty through the posterior distributions of model parameters and of the fairness measure.

Following \citet{stco21}, we define $\Um$ as the estimated residuals from the multivariate ordinary least squares (OLS) regression of $\Xm$ on $\Sm$, $p$ and $q-$dimensional respectively. By definition, $\Um$ represents the part of $\Xm$ which is uncorrelated with $\Sm$. Then, instead of regressing the response on $\Xm,\Sm$, we do it on $\Um,\Sm$. For a continuous response, we define the likelihood as
\begin{align}\label{eq:bayesian_lm}
  y_i \mid b_0,\am,\bm,\sigma^2,\Sm,\Um
  &\overset{\text{ind}}{\sim} \mathcal N\!\big(b_0 + \Sm_i^\top \am + \Um_i^\top \bm,\ \sigma^2\big),
  \qquad i=1,\dots,n.
\end{align}
In this way, the correlation between $\Sm$ and the response appears only as the parameter $\am$, and we can penalise the influence of the sensitive covariates through a fairness penalty $\lambda \ge 0$ on $\am$. We do that through the following prior specification:
\begin{align*}
  \sigma^2 \sim \mathrm{Inv\text{-}Gamma}(a_\sigma, b_\sigma),
  \qquad\boldsymbol{\theta} = (b_0,\am,\bm) \mid \sigma^2,\lambda \sim \mathcal N_{1+q+p}\big(\boldsymbol{\mu}_0,\ \sigma^2 \Lambda_0^{-1}\big),
\end{align*}
where $\mathcal N_{1+q+p}$ is a $(1+q+p)-$variate normal with prior block precision matrix
\begin{align*}
&\Lambda_0 =
\begin{pmatrix}
c & 0 & 0 \\
0 & \lambda\, I_q & 0 \\
0 & 0 & \Um^\top \Um / g
\end{pmatrix},&
&c > 0, g > 0.
\end{align*}
Since the block off-diagonals are zero, this is equivalent to mutually independent Gaussian priors.
The prior for $\bm$ is a Zellner g-prior \citep{zellner1986assessing}: when $g$ approaches infinity, the posterior mean converges to the MLE of the least squares linear regression. The prior for $\am$ is a ridge prior with penalty $\lambda$, which acts as a smooth control of the degree of regularisation on the sensitive coefficients: when $\lambda$ approaches 0, the $\am$ are essentially unconstrained; when $\lambda$ increases, the prior variance decreases, shrinking the $\am$ towards zero and thus reducing the model's dependence on $\Sm$.

In this case, the posterior is available in closed form since the model is conjugate, and estimates can be obtained analytically. Calling $\mathbf{Z}=[\Sm,\, \Um]$ and
\begin{align*}
  \Lambda_n &= \mathbf{Z}^T \mathbf{Z} + \Lambda_0, \qquad\mu_n = \Lambda_n^{-1} (\mathbf{Z}^T  \mathbf{Z} \hat{\boldsymbol{\theta}}_{\text{OLS}} + \Lambda_0  \boldsymbol{\mu}_0), \\
  a_n &= a_\sigma + \frac{n}{2},\qquad b_n = b_\sigma + \frac{1}{2} (y^T y - \boldsymbol{\mu}_n^T \Lambda_n \boldsymbol{\mu}_n + \boldsymbol{\mu}_0^T  \Lambda_0 \boldsymbol{\mu}_0),
\end{align*}
where $\hat{\boldsymbol{\theta}}_{\text{OLS}}$ are the OLS estimates. The posterior distributions are
\begin{align}\label{eq:posterior_lm}
  \sigma^2|y,\Sm,\Um,\lambda&\sim \text{Inv-Gamma}(a_n,b_n)\\
  \boldsymbol{\theta} |y,\Sm,\Um,\lambda,\sigma^2 &\sim N(\boldsymbol{\mu}_n,\sigma^2\Lambda_n^{-1})
\end{align}
As a fairness constraint, we ask for the proportion of the variance of the fitted linear predictor attributable to the sensitive attributes to be capped by the unfairness budget $r$:
\begin{align}\label{eq:bayesian_R}
  R_S^2 =\frac{\text{Var}(\Sm^T\am)}{\text{Var}(\Sm^T\am)+\text{Var}(\Um^T\bm)}\le r.
\end{align}
Indeed, when $r=0$, then $\am=\mathbf{0}$ and the predictor $\hat{\y}$ becomes uncorrelated to $\Sm$. This notion of fairness is equivalent to statistical parity when $(\Sm,\Xm)$ are normally distributed, since this is the only case where uncorrelation implies independence. Indeed, the simulations in Section \ref{sec:simulations} follow this set-up. In general, though, this is not the case and our fairness definition is a weaker version of statistical parity. Due to the difficulty of imposing strict independence, correlation-based measures of fairness are not new to the literature, as seen for example in \cite{zafar}.\\
 
The same construction can be applied to a binary response $Y_i \in \{0,1\}$, replacing the Gaussian likelihood with a Bernoulli:
\begin{align}\label{eq:bayesian_logistic}
  y_i \mid b_0,\am,\bm,\Sm,\Um
    &\overset{\text{ind}}{\sim} \mathrm{Bernoulli}\big(\pi_i\big),
  \qquad
  \pi_i = \mathrm{logit}^{-1}\big(b_0 + \Sm_i^\top \am + \Um_i^\top \bm\big),
  \qquad i=1,\dots,n.
\end{align}
The prior on $\boldsymbol{\theta}$ retains the block structure $\Lambda_0$ introduced above, without the $\sigma^2$ scaling, that is $\boldsymbol{\theta} = (b_0,\am,\bm) \sim \mathcal N_{1+q+p}\big(\mathbf 0,\ \Lambda_0^{-1}\big)$. To set the value of $\lambda$, we use an unfairness measure based on the deviance difference between the fully fair model and the estimated model, relative to the deviance difference between the fully fair model and the fully unfair model:
\begin{equation}\label{eq:bayesian_deviancediff}
r(\am,\bm,\lambda)
  = \frac{D(\bm,\am,\infty) - D(\bm,\am,\lambda)}
         {D(\bm,\am,\infty)-D(\bm,\am,0)}\le r,
\end{equation}
where the deviance $D(\bm,\am,\lambda) = -2 \, \ell(y, \bm,\am,\lambda)$ is defined as twice the negative log-likelihood of the model. This is a legitimate extension of Eq.~\ref{eq:bayesian_R} in the GLM setting: for a Gaussian GLM, the two measures are identical because Eq.~\ref{eq:bayesian_deviancediff} is the residual sum of squares. For more on this measure, see \citet{stco21}. Since posteriors are not available in closed form in the Bernoulli case, we performed inference via Markov Chain Monte Carlo methods, in particular the No-U-Turn Sampler (NUTS) (\cite{hoffman2014no}). As hyperparameters, we use $a_\sigma=2, b_\sigma=1, c=10, g=1000$.\\

In both the Gaussian and the Bernoulli case, we define the value of $\lambda$ with an empirical procedure by finding the value such that the unfairness measure equals the target $r$, plugging in the frequentist estimates of the parameters $\am,\bm$ obtained using the \texttt{fairml} R package by \citet{fairml}. Both measures are in fact monotone functions of $\lambda$ and can be inverted numerically. With $r = 0$ we find a perfectly fair model ($\lambda \to \infty$) and with $r = 1$ an unconstrained one ($\lambda = 0$). When the target $r$ exceeds the unfairness present in the data, then the constraint becomes inactive, and $\lambda$ converges to 0. An ad hoc hyperprior for $\lambda$ can be designed to similarly satisfy the constraint, but we preferred this procedure to easily compare results with the frequentist procedure.

Compared to the bootstrap-based approaches described in the previous subsection, the Bayesian construction allows for a more thorough evaluation of the trade-off between model fairness and accuracy by enabling the study of various posteriors. Indeed, since both fairness measures $r(\am,\bm,\lambda)$ and $R_S^2$ are functions of the model parameters, their distributions can be studied by drawing from the parameters' posteriors. Inspecting the unfairness a posteriori informs us on the variability of the model's estimates under the data we have observed: a large variance in the unfairness measure might suggest harsh discrepancies between prior and data and that the data do not support well the required level of fairness. In contrast, a concentrated posterior could reassure the analyst that the target fairness will be achieved with a set confidence. Finally, studying the posterior distributions for different values of $r$ can suggest how to set $r$ to achieve the level of fairness with a certain confidence level.

\section{Results}
\label{sec:results}

\subsection{Simulated data}
\label{sec:simulations}

We assess the approaches described in Section~\ref{sec:uq} on simulated data. In both settings, we sample a design matrix $[\Sm,\Xm]$ from a multivariate normal distribution $\mathcal{N}(\mathbf{0}, \Sigma)$, with $\Sigma$ designed to achieve a predetermined level of inter-correlation $\text{corr}(S,X)$ and of intra correlation in $\Sm$ and $\Xm$. The response depends linearly on all variables: it is Gaussian for linear regression $$y\sim N(\Sm^T\tilde{\am} + \Xm^T\bm, \sigma^2),\quad\sigma=0.1,$$ and Bernoulli for logistic regression $$y\sim \text{Bernoulli}(\text{logit}^{-1}(\Sm^T\tilde{\am} + \Xm^T\bm)).$$ Note the use of $\tilde{\am}$ instead of $\am$: the likelihood of our models in Eq.~\ref{eq:bayesian_lm} and~\ref{eq:bayesian_logistic} do not match the data generating processes (the transformation being $\am=\tilde{\am}+\Gamma^T\bm$, where $\Gamma$ is the matrix of regression coefficients of $\Xm$ over $\Sm$).

\subsection{Frequentist: the bootstrap}

Data generation was completed using the coefficients $\tilde{\am}=[5,6,7],\,\bm=[2,3,4]$,
with correlation of 0.8 within each group of covariates and $\text{corr}(S_i,X_j) \in\{ 0.1, 0.2, 0.5\}$ for all $i=1,\dots,3,\,j=1,\dots,3$. We consider the sample sizes $n \in\{ 50, 100, 200, 500\}$ and the target unfairness levels $r \in\{ 0.05, 0.1, 0.2\}$. We replicate each combination 50 times, and we estimate all models with the \texttt{fairml} package \citep{fairml}. We take $B_1 = 100$ first-level and $B_2 = 20$ second-level bootstrap samples (50 for logistic regression). Interval lengths are relative to a reference interval obtained by refitting the model on 100 independently generated samples. We report all intervals at the 95\% level, averaged over the three sensitive attributes.\\

\begin{figure}
  \centering
  \includegraphics[width=0.32\linewidth, alt={Three bar plots representing the probability of inclusion of true parameters in the confidence intervals for the sensitive coefficients in linear regression generated using nonparametric, double and residual bootstrap. From left to right, the plots represents the probability against sample size, unfairness level and correlation between S and X.}]{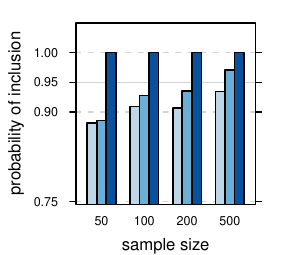}
  \includegraphics[width=0.32\linewidth]{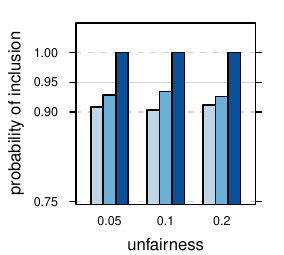}
  \includegraphics[width=0.32\linewidth]{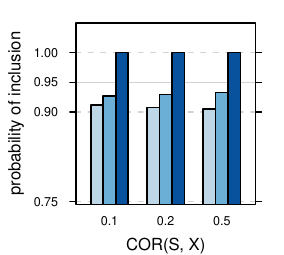}
  \caption{Probabilities of inclusion of the asymptotic estimates of the $\am$ in the 95\% confidence intervals for nonparametric (light), double (mid) and residual bootstrap (dark blue), for linear regression. Each panel varies one simulation parameter and averages over the others: sample size (left), target unfairness $r$ (centre) and $\text{corr}(S, X)$ (right).}
\label{fig:lin-inc-prob}
\end{figure}

\begin{figure}
  \centering
  \includegraphics[width=0.32\linewidth, alt={Three bar plots representing the lengths of confidence intervals of the sensitive coefficients in linear regression (relative to the reference interval) generated using nonparametric, double and residual bootstrap. From left to right, the plots represents the probability against sample size, unfairness level and correlation between S and X.}]{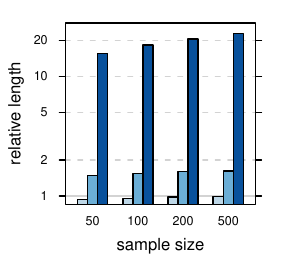}
  \includegraphics[width=0.32\linewidth]{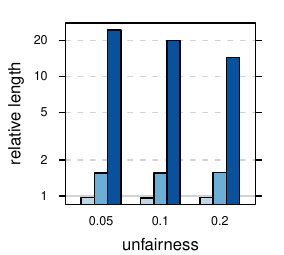}
  \includegraphics[width=0.32\linewidth]{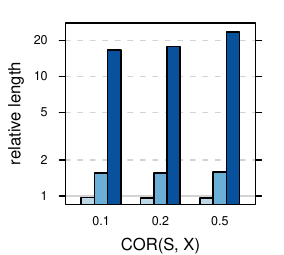}
  \caption{Length of the confidence intervals of the $\am$ for nonparametric (light), double (mid) and residual bootstrap (dark blue) for linear regression. Lengths are relative to the reference interval, so that a value of 1 means the method reproduces it; the vertical axis is on a $\log_{10}$ scale. Each panel varies one simulation parameter and averages over the others: sample size (left), target unfairness $r$ (centre) and $\text{corr}(S, X)$ (right). Geometric mean over the three $\am$.}
\label{fig:lin-int-length}
\end{figure}

\begin{figure}
  \centering
  \includegraphics[width=0.32\linewidth, alt={Three bar plots representing the probability of inclusion of true parameters in the confidence intervals for the sensitive coefficients in logistic regression generated using nonparametric and double bootstrap. From left to right, the plots represents the probability against sample size, unfairness level and correlation between S and X.}]{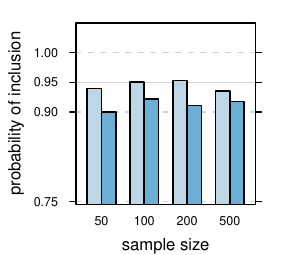}
  \includegraphics[width=0.32\linewidth]{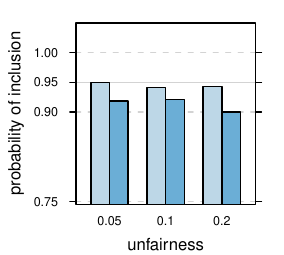}
  \includegraphics[width=0.32\linewidth]{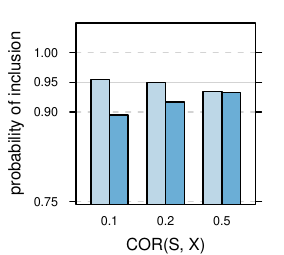}
  \caption{Probabilities of inclusion of the asymptotic estimates of the $\am$ in the 95\% confidence intervals for nonparametric (light) and double bootstrap (mid blue), for logistic regression. Each panel varies one simulation parameter and averages over the others: sample size (left), target unfairness $r$ (centre) and $\text{corr}(S, X)$ (right). Median over the three $\am$.}
\label{fig:log-inc-prob}
\end{figure}

\begin{figure}
  \centering
  \includegraphics[width=0.32\linewidth, alt={Three bar plots representing the lengths of confidence intervals of the sensitive coefficients in logistic regression (relative to the reference interval) generated using nonparametric and double bootstrap. From left to right, the plots represents the probability against sample size, unfairness level and correlation between S and X.}]{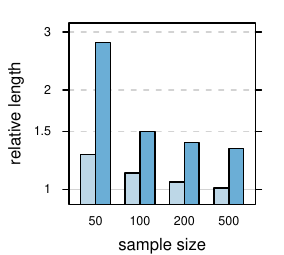}
  \includegraphics[width=0.32\linewidth]{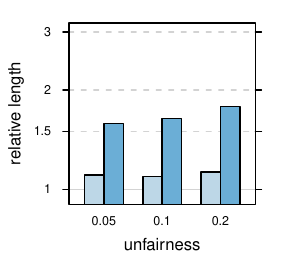}
  \includegraphics[width=0.32\linewidth]{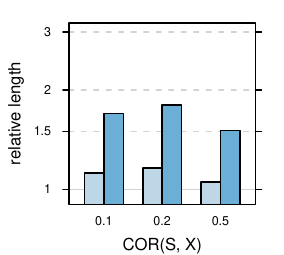}
  \caption{Length of the confidence intervals of the $\am$ for nonparametric (light) and double bootstrap (mid blue) for logistic regression. Lengths are relative to the reference interval, so that a value of 1 means the method reproduces it; the vertical axis is on a $\log_{10}$ scale. Each panel varies one simulation parameter and averages over the others: sample size (left), target unfairness $r$ (centre) and $\text{corr}(S, X)$ (right). Geometric mean over the three $\am$.}
\label{fig:log-int-length}
\end{figure}

Figure~\ref{fig:lin-inc-prob} shows that for $\am$ double bootstrap has higher inclusion probabilities than nonparametric bootstrap in every configuration we considered in the gaussian case. Nonparametric bootstrap undercovers systematically (0.89--0.94), while double bootstrap improves on it throughout and shows mild over-coverage between $n \in \{ 200, 500\}$, where it reaches 0.97. Averaged over the simulation grid, double bootstrap is closer to the nominal level, although at $n = 500$ its overshoot is marginally larger than the shortfall of nonparametric bootstrap. Residual bootstrap has an empirical coverage of exactly one at all sample sizes and for all values of $r$ and of the correlation between $\Sm$ and $\Xm$. For the other methods, coverage improves with the sample size and is not affected by either $r$ or the correlation between $\Sm$ and $\Xm$.

Figure~\ref{fig:lin-int-length} shows the length of different confidence intervals for $\am$ relative to those computed directly from the data generating process. Those from the nonparametric bootstrap have almost exactly the same length; their undercoverage arises from location rather than interval width. Double bootstrap produces intervals about 60\% longer, achieving higher coverage. Residuals bootstrap, on the other hand, produces intervals about 20 times longer than the reference, making it useless in practice. Overall, inflation grows as $r$ decreases, from 15 ($r = 0.2$) to 25 ($r = 0.05$). It also grows with the correlation between $\Sm$ and $\Xm$ and with the sample size. Smaller values of $r$ correspond to larger penalties: as mentioned in Section~\ref{sec:uq}, the empirical residuals become less and less pivotal as the fairness penalty shrinks the coefficients, and the problem does not disappear as $n$ grows.

For logistic regression, only nonparametric and double bootstrap are applicable. In Figure~\ref{fig:log-inc-prob}, their coverage ranking is reversed: nonparametric bootstrap is always nominal (0.94--0.96), while double bootstrap always undercovers (0.90--0.93). Figure~\ref{fig:log-int-length} shows that nonparametric bootstrap intervals are close in length to the reference, while those from double bootstrap are always longer, increasingly so for smaller sample sizes. Therefore, nonparametric bootstrap is preferable here on both counts, which is the opposite of what we found for linear regression.

\subsection{Bayesian: studies a posteriori}

The configuration of parameters used in this section is the following: $\tilde{\am}=[1,2],\,\bm=[3,4]$, $\text{corr}(S_i, X_j) \in\{ 0.1, 0.3\}$, $r \in\{0.05, 0.1\}$, $n \in\{100, 500\}$. We present the results for the linear regression case, since the results and considerations for logistic regression are similar.

\begin{figure}
  \centering
  \includegraphics[width=0.8\linewidth, alt={A two-by-two grid of violin plots showing the posterior distribution of the fairness measure $R^2_S$. Rows show results at target unfairness $r = 0.05$ and $r = 0.10$; columns are correlation between S and X of 0.1 and 0.3. Each panel shows violins for sample sizes
  100 and 500, with a red cross marking the posterior mean.}]{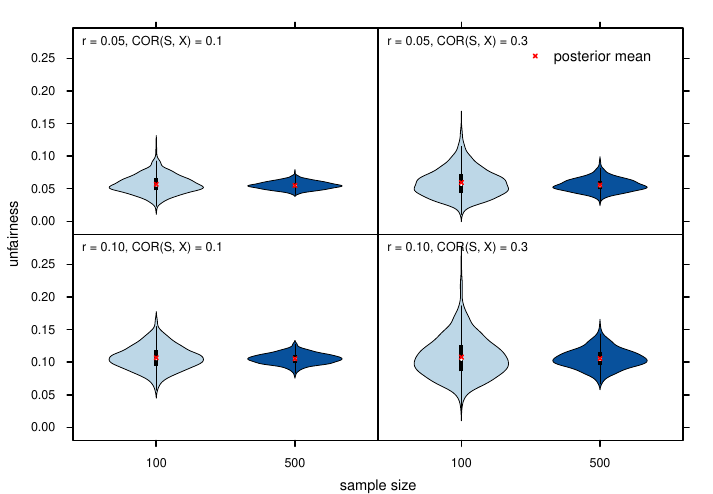}
  \caption{Violin plot of the empirical posterior distribution of $R_S^2$. Along the columns, different values of unfairness $r$, and along the rows the correlation $\text{corr}(\Sm,\Xm)$. Each plot reports two sample sizes.}
\label{fig:R2_posterior}
\end{figure}

\begin{figure}
    \centering
    \includegraphics[width=.8\linewidth,alt={A two-by-two grid of scatter plots of rescaled unfairness against accuracy. Rows are sample sizes 100 and 500; columns are correlation between S and X of 0.1 and 0.3. Each point is one simulated data set, coloured by its target unfairness r.}]{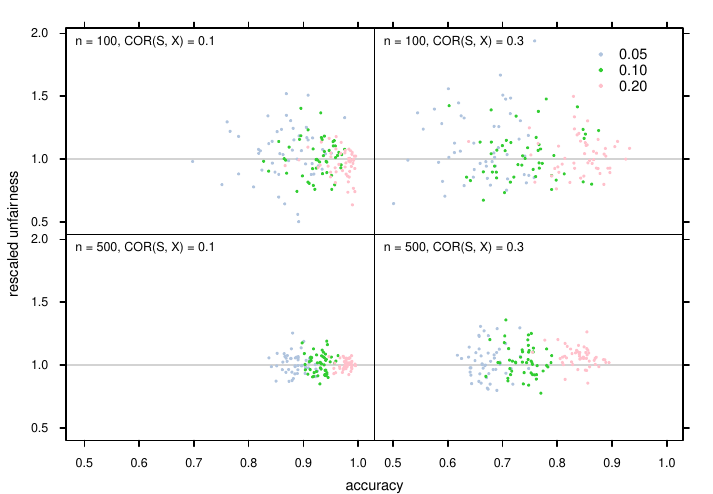}
    \caption{Accuracy (measured as coefficient of determination) vs rescaled unfairness. Along the columns, different sample sizes, and along the rows, different levels of correlation $\text{corr}(\Sm,\Xm)$. Each dot represents a simulated data set, and the colour identifies the unfairness value $r$.}
    \label{fig:unfairness_vs_accuracy}
\end{figure}

A crucial difference with the previous section is that the Bayesian procedure allows us to study the posterior distribution of the unfairness measure directly: for each set-up, we sampled 200 values from the posterior of $\am,\bm,\sigma^2$ and the corresponding values of $R_S^2$ (Eq.~\ref{eq:bayesian_R}), which we used to produce the violin plots in Figure~\ref{fig:R2_posterior}. The posterior mean is centred slightly above the user-defined level of $r$. This is because the empirical mean of a non-linear functional is a biased estimate due to Jensen's inequality: the fairness measure is convex in the parameters, so its empirical mean exceeds its value at the posterior mean. Since the constraint is applied to the point estimates, the unfairness measured on those is correct (as we will illustrate in Figure~\ref{fig:unfairness_vs_accuracy}). As expected, the posterior variance decreases as the sample size increases. More interestingly, for the same $r$, the variance of $R_S^2$ increases as the correlation between $\Xm$ and $\Sm$ increases. We hypothesise that, as $\text{corr}(\Sm,\Xm)$ grows, $\Sm$ becomes an increasingly effective proxy for the legitimate covariates $\Xm$, so a larger share of the outcome's predictive signal passes through $\Sm$ and less from $\Um$. 
The constraint $R^2_S = r$ nevertheless caps the proportion of explained variance attributable to $\Sm$s. The fitted model is therefore obliged to discard progressively more signal, which re-emerges in the residual variance $\sigma^2$, which contributes to increasing the variance of $R_S^2$. We also notice that the posterior is asymmetric, skewed toward more unfairness than required.

In Figure~\ref{fig:unfairness_vs_accuracy} we investigate the behaviour of the posterior mean of $R_S^2$ under data variability against the accuracy of the estimation, measured by the coefficient of determination of the regression. To generate data, we repeat the illustrated data generation 50 times for each combination of correlation, unfairness and sample size, and for each data set we computed the unfairness on the posterior mean of $\am,\bm$, rescaling it by dividing by the corresponding $r$. For each scenario, the rescaled measure is centred on 1, meaning our procedure targets the correct value. On the accuracy side, as the unfairness constraint loosens, accuracy increases, which is a well-known and expected trade-off. Moving along the rows (same $n$ and $r$), the clouds of dots move left toward lower accuracy because the predictors are more correlated at the same level of fairness.
For $n=100,\text{corr}(\Sm, \Xm)=0.1$, we observe that some of the data sets with constraint $r=0.2$ achieve maximum accuracy and rescaled unfairness less than 1: for those data, the requested $r$ was lower than the observed unfairness of the data, making the constraint inactive and the regression problem fall back to the OLS framework.

Our considerations suggest caution: although the model is guaranteed to provide fair predictors at the point estimates, the uncertainty in the fairness measure can be quite high, with the posterior reaching less fair values than requested with probability greater than 0.5. As the correlation between $\Sm$ and $\Xm$ increases, this problem becomes even more evident, and a non-null mass of probability lies on values of $R^2_S$ that can even be $50\%$ more unfair than the target.\\

\begin{figure}
    \centering
    \includegraphics[width=0.8\linewidth, alt={A two-by-two grid of profile plots against target unfairness r. The top row shows the two sensitive coefficients alpha, the bottom row the two legitimate coefficients beta; columns are correlation between S and X of 0.1 and 0.3. Each series is a line with a shaded 95\% credible band and error bars.}]{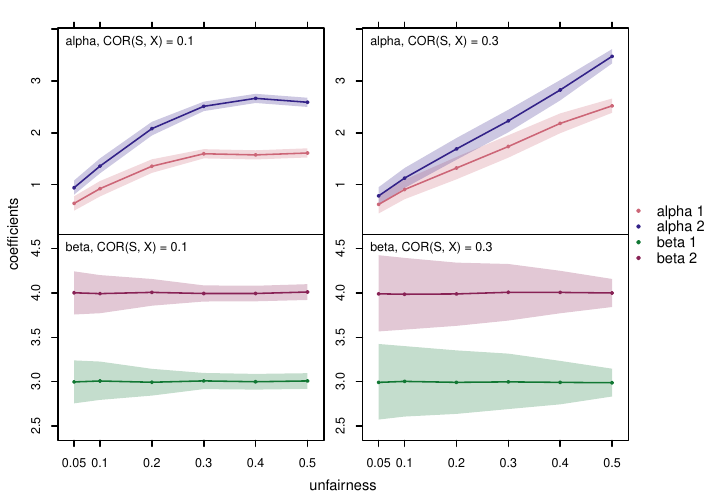}
    \caption{Profile plots of the posterior means of $\am$ (top) and $\bm$ (bottom) and their 95\% credible intervals, averaged over 50 simulated data sets with different levels of correlation between $\Sm,\Xm$ (0.1 on the left, 0.3 on the right).}
    \label{fig:profile_plots_bayescoef}
\end{figure}

In Figure~\ref{fig:profile_plots_bayescoef} we show the profile plots of the estimates of $\am,\bm$ at different levels of $r\in\{0.05, 0.1, 0.2, 0.3, 0.4, 0.5\}$, averaged over 50 data sets with sample size 500. We observe that for $\text{corr}(\Sm,\Xm)=0.1$, the original data sets have a value of $R_S^2$ close to 0.3, which means that when $r$ reaches and exceeds that level, the constraint becomes inactive, $\lambda$ goes to zero, and the estimated coefficients go to the OLS estimates. This does not happen for $\text{corr}(\Sm, \Xm)=0.3$, since the data $R_S^2$ is higher than 0.5. As expected given our modelling choices, the posterior mean of $\bm$ is always centred on the true values, while $\am$ starts near 0 (when the penalisation is high) and increases until the OLS regime is reached. The credible intervals of $\bm$ shrink as $r$ increases and then stabilise when the constraint becomes inactive. This is an expected outcome of Eq.~\ref{eq:posterior_lm}, where the posterior of $\sigma^2$ depends on $\lambda$ even for $\bm$. Uncertainty for $\am$ is instead more stable.

\subsection{Real data}
\label{sec:real_data}

We illustrate and compare our proposals on two widely-used benchmark data sets from the fairness literature, both available in the \texttt{fairml} R package \citep{stco21}. The law school admissions data \citep[LSAC;][]{sander} records the careers of 20,800 students surveyed in 1991; we model the undergraduate GPA with a linear regression. The COMPAS data \citep{propublica} features 5,855 defendants, and we model their recidivism within two years with a logistic regression. In both cases, race, sex and age are the sensitive attributes, and the remaining variables are the predictors (12 for COMPAS, 11 for LSAC); we let the target unfairness $r$ vary between 0.01 and 0.5. For the frequentist approach, we stratify resampling for COMPAS jointly by race and sex because some combinations are rare: the smallest comprise two observations. The frequentist model is the one proposed by \citet{stco21}. In this way, the frequentist and Bayesian estimates of the models' coefficients are guaranteed to be the same (up to numerical error).\\

\begin{figure}
  \centering
  \includegraphics[width=0.32\linewidth, alt={Line plots of the confidence or credible intervals and point estimates of the six sensitive coefficients for the LSAC data against target unfairness r, on a logarithmic horizontal axis from 0.01 to 0.5. Each series has a shaded 95\% credible band and error bars. From left to right: nonparametric, double bootstrap and posterior credible intervals.}]{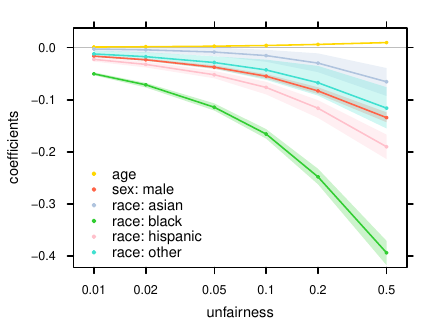}%
  \hfill%
  \includegraphics[width=0.32\linewidth]{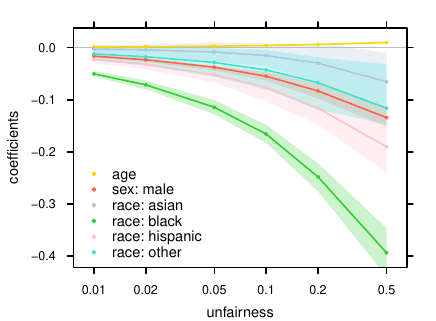}
  \includegraphics[width=0.32\linewidth]{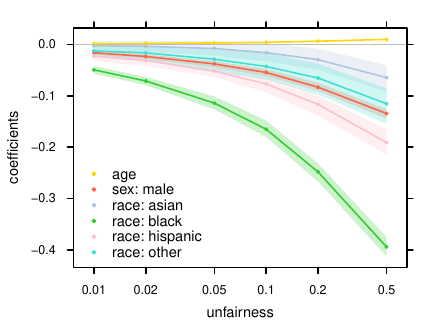}
  \caption{LSAC. Confidence and credible intervals (95\%) for the sensitive attributes estimated using nonparametric (left), double bootstrap (centre) and the Bayesian model (right).}
\label{fig:lsac-boot}
\end{figure}

\begin{figure}
  \centering
  \includegraphics[width=0.32\linewidth, alt={Line plots of the confidence or credible intervals and point estimates of the six sensitive coefficients for the COMPAS data against target unfairness r, on a logarithmic horizontal axis from 0.01 to 0.5. Each series has a shaded 95\% credible band and error bars. From left to right: nonparametric, double bootstrap and posterior credible intervals.}]{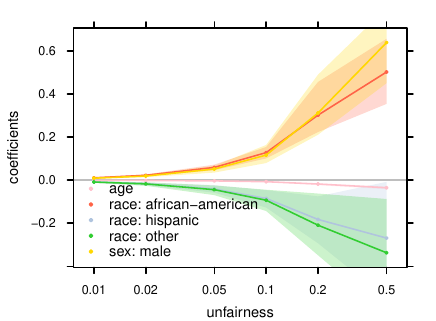}
  \includegraphics[width=0.32\linewidth]{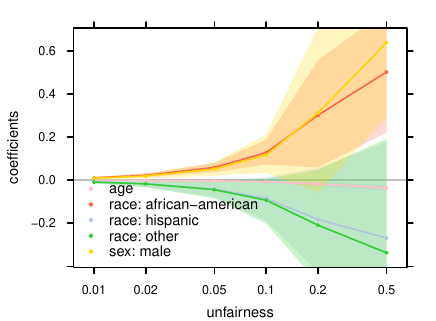}
  \includegraphics[width=0.32\linewidth]{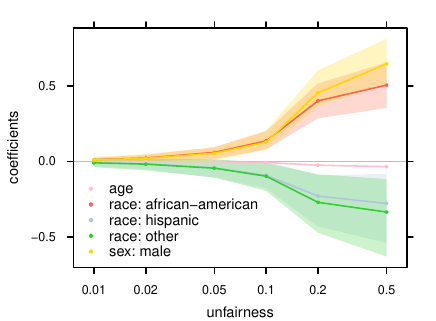}
  \caption{COMPAS. Confidence and credible intervals (95\%) for the sensitive attributes estimated using nonparametric (left), double bootstrap (centre) and the Bayesian procedure (right).}
\label{fig:compas-boot}
\end{figure}

We choose the value of $r$ rather than estimating it, so it has no associated sampling or prior distribution. Nevertheless, its effect can still be assessed. Point estimates alone differ for every $r$ and do not allow us to establish whether two values of $r$ produce statistically significant models, but if the intervals estimated for the $\am$ and $\bm$ at $r$ and at a nearby value have a statistically significant overlap, the two corresponding models are not distinguishable. LSAC and COMPAS differ in this respect. LSAC's bootstrap confidence and credible intervals in Figure~\ref{fig:lsac-boot} are narrow enough to distinguish models estimated at neighbouring targets. COMPAS's bootstrap confidence and credible intervals in Figure~\ref{fig:compas-boot} are not (visually, this is clear for $r > 0.1$): the models at $0.2$ and at $0.5$ overlap, even though their point estimates are far apart. In the same figures, the plots on the right present the posterior means and credible intervals derived from the Bayesian procedure, which mostly agree with the nonparametric bootstrap.

At a given level of fairness, both the confidence and credible intervals in Figures~\ref{fig:lsac-boot} and~\ref{fig:compas-boot} tell us which sensitive attributes have an effect that is distinguishable from zero. However, we cannot use the same intervals to choose $r$: tightening the constraint until the coefficients of the sensitive attributes stop being significant does not work in general because the intervals may shrink at least as fast as the coefficients. Intervals help us interpret a model at a given level of fairness; they do not help us select that level.
Which effects we call significant also depends on how we quantify the uncertainty. Nonparametric and double bootstrap disagree on the same data and the same model, and the credible intervals give a third answer. The frequentist and Bayesian analyses also answer different questions.\\

\begin{figure}
  \centering
  \includegraphics[width=0.45\linewidth, alt={On the left, a scatter plot of rescaled unfairness against accuracy for the LSAC data, bootstrapping over the data set. On the right, six points, one per target unfairness r, each with horizontal and vertical 95\% credible intervals and a shaded covariance ellipse for the posterior of the rescaled unfairness against accuracy. Colors identify the level of r from 0.01 to 0.5.}]{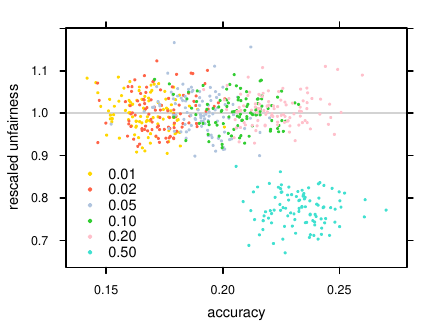}%
  \includegraphics[width=0.45\linewidth]{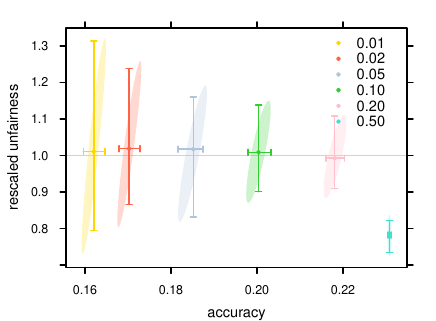}
  
  \caption{LSAC. Marginal empirical fairness against accuracy (coefficient of determination $R^2$). The empirical fairness values are divided by the target $r$, so that 1 denotes exact attainment. On the left, the graph is obtained by bootstrap resampling. On the right, the posterior credible intervals and the ellipse showing the covariance between accuracy and unfairness.}
\label{fig:lsac-risk}
\end{figure}

\begin{figure}
  \centering
  \includegraphics[width=0.45\linewidth, alt={On the left, a scatter plot of rescaled unfairness against accuracy for the COMPAS data, bootstrapping over the data set. On the right, six points, one per target unfairness r, each with horizontal and vertical 95\% credible intervals and a shaded covariance ellipse for the posterior of the rescaled unfairness against accuracy. Colors identify the level of r from 0.01 to 0.5.}]{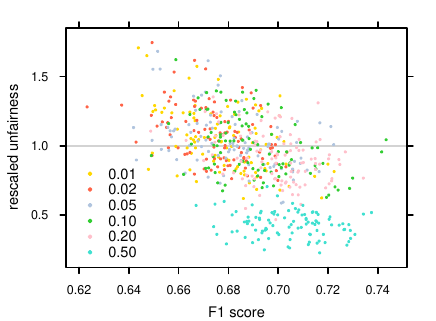}%
  \includegraphics[width=0.45\linewidth]{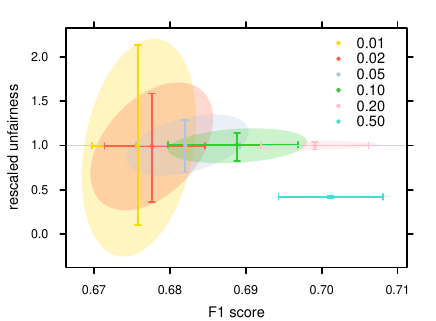}
  \caption{COMPAS. Marginal empirical fairness against accuracy (F1 score). The empirical fairness values are divided by the target $r$, so that 1 denotes exact attainment. On the left, the graph is obtained by bootstrap resampling. On the right, the posterior credible intervals and the ellipse showing the covariance between accuracy and unfairness.}
\label{fig:compas-risk}
\end{figure}

Figures~\ref{fig:lsac-risk} and~\ref{fig:compas-risk} describe the spread of the fairness level over hypothetical replications of the data (left) and the posterior distribution given the sample we observed (right, in the form of credible intervals) versus accuracy. Only the former involves the variability of the data, as we noted in Section~\ref{sec:uncertainty}. The plots for the two data sets are not directly comparable: the fairness and accuracy metrics are different. 

The bootstrap shows that constraining a model to a target $r$ produces a distribution of realised unfairness centred close to $r$, with part of it lying above it. 
In the Bayesian framework, resampling the data would not make sense; hence, we plot the posterior spread of the same quantities, computed over 200 samples. Once again, due to Jensen's inequality, the posterior of the unfairness is slightly above the target and, similarly to the simulations, as $r$ increases, the unfairness' posterior variance decreases and the accuracy increases. The ellipses show the covariance of the bivariate distribution: they tilt toward the right due to the trade-off between fairness and accuracy; less fair posterior draws are more accurate. Interestingly, the deviance ratio for the Bayesian predictors might not be positive, since that is guaranteed only for the MLE estimator. 
The Bayesian version gives another point of view on the impact of the choice of $r$: while LSAC gives different responses for different values of $r$, the COMPAS models overlap. The latter plot suggests caution with low values of $r$: the credible intervals for unfairness suggest variability that is too large to be effective, indicating that the data do not support a strict constraint. In both data sets, the unfairness level supported by the data lies between 0.2 and 0.5, where the constraint becomes inactive and the fairness of the model depends only on the data.

Together, these plots show that compliance is a probabilistic statement, and we can only make it if we quantify uncertainty in the fairness level. To meet a threshold with high probability, we must set the target below it, and how far below depends on the data. The spread around the target is wider for COMPAS than for LSAC, so the same target leaves a different probability of exceeding the threshold.

\section{Conclusions}

The algorithmic fairness literature usually formulates the learning problem as a constrained optimisation with a single solution, and reports both the model that solves it and the fairness level it achieves as point estimates. In our paper, we argue that this is not a safe procedure for assessing models that are put to use in sensitive settings that come with moral and legal obligations.
In this work, we have limited ourselves to in-processing methods, since post-processing cannot account for the uncertainty of selecting the model and estimating its parameters, and pre-processing increases uncertainty. Within that class, we illustrated one frequentist and one Bayesian approach; many others would have served equally well. The frequentist approach describes how the fairness level varies over repeated samples; the Bayesian one describes how it is distributed given the sample we have. A regulator asking whether a procedure usually produces compliant models needs the first; an analyst asking whether this model complies needs the second. Neither question is more legitimate than the other, and neither substitutes for the other.

The legal frameworks that govern AI systems require a thorough risk assessment of the models they use to identify and contain the biases that may make the model unfair within acceptable, practical limits. As we have shown with both simulations and real data, a model constrained to a target level of unfairness
does not necessarily attain that level. It attains a distribution of levels, centred near
the target when the estimator is well behaved; part of that distribution
will inevitably be higher than the target, making the predictions potentially unfair when we consider the intrinsic variability of the data generating process and that of model selection and estimation. Whether a model complies is therefore
a probabilistic statement, one which we cannot make from a point estimate. At the very least, we should report an interval or a posterior distribution. The NIST AI Risk Management Framework effectively requires us to do that: it defines risk as ``the composite measure of an event's probability of occurring and the magnitude or degree of the consequences of the corresponding event'', and asks that performance be assessed ``with associated measures of uncertainty'' \citep{nist-ai}.

Two practical consequences follow. To meet a threshold with high probability, we
must set the target below it to bound the probability of exceeding it, which represents the risk of noncompliance. How far below depends on the data but can be quantified, as we demonstrated with the LSAC and COMPAS data sets. The
target itself is chosen as a policy decision rather than estimated, so it carries no uncertainty of
its own. The only way to judge how sensitive our choices are to its value is
to look at the uncertainty of everything else, including model parameters and unfairness measures. Because the fairness measure is a function of the parameters, Bayesian methods give us a clear idea of how far we might get from our target by plotting its posterior distribution. A sensitivity analysis on the values of unfairness may find out that we are not able to
distinguish models estimated at similar target values. It also helps us discern the implications of our choices, for example suggesting that, as in the COMPAS example, certain values of $r$ are unlikely to produce the desired fairness, and with such an uncertainty we might prefer trading off some fairness and gain instead on accuracy, or evaluate different notions of fairness and their implications.

Uncertainty quantification must itself be done with care, because the available
methods may provide different answers. In our simulations, double bootstrap gave the coverage closest to nominal for linear regression, whereas resampling the residuals was unusable because the confidence intervals it produced were an order of magnitude too long. For logistic regression, the ranking of the first two methods reversed. No method dominated in both settings, suggesting we should choose the method that fits the specific setting at hand. Bayesian credible intervals give answers similar to the nonparametric intervals, suggesting possibly better coverage for logistic rather than linear regression. Overall, we find that the penalty enforcing fairness in a model may reduce uncertainty in estimating some of its parameters, but the choice of the penalty itself carries its own uncertainty. The overall uncertainty shifts from model estimation to model selection instead of decreasing.

Our analyses represent a preliminary step in the task of risk assessment for fair AI systems. Multiple directions can be taken to extend what has been done so far. Our Bayesian proposal matches the model of \cite{stco21}, but it could be made a stand-alone, fully Bayesian model without relying on frequentist estimates to select $\lambda$. More generally, Bayesian approaches have only recently started to prove their usefulness in fair ML \citep{chiappa2018causal, dimitrakakis2019bayesian,  NEURIPS2020_d83de59e, tahir2023fairness, carrizosa2026empirical}, but they are still rare in the field despite carrying a natural notion of uncertainty: the model's posterior distribution can indeed be used to assess the fairness of a predictive problem more thoroughly than other methods.
With respect to the literature on counterfactual and path-specific fairness \citep{chiappa2018causal, kusner2017counterfactual}, our approach is agnostic to which paths are allowed to transport sensitive information, assuming all paths are not admissible. Future work could design methods to quantify uncertainty also in causal settings.



\section{Data availability}
The data for Section~\ref{sec:simulations} are simulated. The data for Section~\ref{sec:real_data} are open source and can be found in the \texttt{fairml} R package by \citet{fairml} or in the code reporsitory. The code is attached to the submission and will be published in an open-source online repository.
 

\section{Author contributions statement}

The authors contributed equally to the design and conceptual work of the paper. F.P. wrote the code for the Bayesian parts, M.S. for the frequentist part. F.P. and M.S. jointly wrote the paper.

\section{Acknowledgments}

During the preparation of this work, the authors used Anthropic's Claude Opus 5 to revise and rewrite the Python code in the replication repository, to reformat the figures, for copy-editing of the manuscript and for exploratory analysis that informed the discussion of the posterior behaviour of the fairness measure in Sections \ref{sec:simulations} and \ref{sec:real_data}. The authors reviewed all the material, and take full responsibility for the content of the publication.

\section{Funding}
F.P. thanks the Department of Informatics of Università della Svizzera Italiana, which granted her the 2025 Visiting Lecturer position to work on this project.

\bibliographystyle{abbrvnat}
\bibliography{reference}

\end{document}